# Multi-level decision framework collision avoidance algorithm in emergency scenarios

Guoying Chen[1], Xinyu Wang[1], Min Hua[2*], Wei Liu[3]
[1]State Key Laboratory of Automotive Simulation and Control, Jilin University, Changchun, Jilin, China
[2]School of Engineering, University of Birmingham, Birmingham, UK
[3]School of Electrical and Computer Engineering, Purdue University, West Lafayette, US
*Corresponding author: Min Hua (e-mail: mxh623@student.bham.ac.uk).

**Abstract:** With the rapid development of autonomous driving, the attention of academia has increasingly focused on the development of anti-collision systems in emergency scenarios, which have a crucial impact on driving safety. While numerous anti-collision strategies have emerged in recent years, most of them only consider steering or braking. The dynamic and complex nature of the driving environment presents a challenge to developing robust collision avoidance algorithms in emergency scenarios. To address the complex, dynamic obstacle scene and improve lateral manoeuvrability, this paper establishes a multi-level decision-making obstacle avoidance framework that employs the safe distance model and integrates emergency steering and emergency braking to complete the obstacle avoidance process. This approach helps avoid the high-risk situation of vehicle instability that can result from the separation of steering and braking actions. In the emergency steering algorithm, we define the collision hazard moment and propose a multi-constraint dynamic collision avoidance planning method that considers the driving area. Simulation results demonstrate that the decision-making collision avoidance logic can be applied to dynamic collision avoidance scenarios in complex traffic situations, effectively completing the obstacle avoidance task in emergency scenarios and improving the safety of autonomous driving.

## 1 Introduction

As the global economy continues to grow at a rapid pace, the rate of car ownership has been on the rise. The World Health Organization (WHO) reports that road traffic accidents cause nearly 1.3 million deaths and approximately 50 million injuries worldwide each year and are the leading cause of death among children and young people worldwide. The WHO released "Global Plan Decade of Action for Road Safety at its headquarters in Geneva, Switzerland", calling on countries to take measures to reduce road traffic fatalities and injuries by at least 50% by 2030 at the latest (WHO, 2021). According to the data, about 92% of road traffic accidents are due to improper operation of drivers. In an emergency situation, drivers are not aware of the risk of encountering or have too little time to react to take the right steps to avoid a collision (Liu et al., 2023; Xia et al., 2022).

The industry commonly uses passive safety control systems to improve driving safety, including an anti-lock brake system (ABS), traction control system (TCS), electronic stability controller (ESC), etc (Bengler et al., 2014; Chen et al., 2023). Although passive safety systems are crucial in reducing the severity of vehicle collisions, the continuous development of sensors and data processing technologies has led to the emergence of advanced driving assistance systems (ADAS), which can detect potential dangers in various scenarios and make prompt decisions (Liu et al., 2022). By detecting potential hazards and making correct decisions in different scenarios, ADAS systems can significantly enhance driving safety (Xiong et al., 2020; Hua et al., 2019). In case of emergency, the active collision avoidance movement can be divided into two major categories, i.e., collision avoidance by changing the longitudinal motion of the vehicle through emergency braking and collision avoidance by steering or braking to make the vehicle change lanes in the form of emergency lane change to avoid the obstacle (Gao et al., 2022; Hua et al., 2020; Liu et al., 2021). Longitudinal collision avoidance technologies, represented by Forward Collision Warning (FCW) systems and Automatic Emergency Braking (AEB) systems, warn the driver when there is a risk of collision in front of the vehicle or automatically apply the braking operation when an accident is about to occur, and such systems These systems have been commercially implemented in many vehicles and have effectively improved vehicle safety (Cicchino et al., 2017; Chen et al., 2019). However, when the vehicle is in emergency conditions such as low road adhesion coefficient, high relative speed to obstacles, or oncoming traffic in the opposite direction, the impending collision cannot be avoided by longitudinal braking, and then the vehicle can avoid collision by emergency steering around the obstacles.

Han et al. proposed a practical probabilistic method for collision decision-making, where a Gaussian hybrid method is designed to calculate the collision probability with the help of linear recursive collision time (TTC) estimation (Han et al., 2015). Experiments show that the method can greatly reduce the inherent nonlinearity of the collision decision problem and the complexity of the collision probability calculation. The method improves the reliability of the collision probability calculation and provides a solution for developing real-time decision algorithms. Huang et al. proposed a novel bi-exponential TTC decision algorithm that can distinguish between safe passage in adjacent lanes and basic hazard lateral collision situations (Huang et al., 2010). In addition, gray prediction theory is introduced to estimate the relative distance of two vehicles one step ahead, and this strategy can effectively avoid collisions. The results showed that with the reduction of emergency braking system delay and collision time trigger threshold increases, the percentage of pedestrian collisions avoided also increases, and several studies have highlighted the importance of low system delays for vehicle-equipped autonomous braking systems and that choosing the appropriate crash time threshold for emergency braking systems is key to determining system usability, with too low a threshold leading to an overly sensitive system and resulting in false alarms, and too high a threshold increasing the risk of vehicle crashes (Haus et al., 2019). In order to avoid collisions between vehicles and pedestrians while crossing the road, Yang et al. established an AEB-P warning model based on TTC (time to collision) and braking safety distance, and defined the traffic safety level and working area of the AEB-B warning system, and verified the plausibility of this control strategy through experiments (Yang et al., 2019). Tian et al. proposed a control strategy based on fuzzy logic control for the crash risk assessment model and corresponding pedestrian AEB control strategy (Tian et al., 2022). The results show that the developed AEB control strategy can accurately assess the collision risk and take effective measures to avoid collisions with pedestrians crossing the road at a constant speed. Considering the accuracy and timeliness of automatic system control, Hang et al. proposed a rear-end real-time automatic emergency braking (RTAEB) system (Hang et al., 2022). Real-time driver-based conflict recognition and collision avoidance performance are inserted for braking intervention. The results show that the system can help to successfully avoid all collision events, and the TTC threshold of 1.5 s and the maximum deceleration threshold of -7.5 m/s$^2$ can achieve the best collision avoidance effect.

In an emergency situation, when the safety distance is short, the vehicle can quickly employ steering to perform collision avoidance operations. Lane changes are limited by friction between the tires and the road surface, which imposes coupled limits on the lateral and longitudinal acceleration of the vehicle, making steering operations more difficult (Chen et al., 2019; Xia et al., 2022; Zhou et al., 2023). Funke et al. demonstrated that self-driving cars can perform emergency lane changes in the friction limit by generating and evaluating binary paths in real-time (Funke et al., 2016). Vehicle emergency active collision avoidance path planning is one of the important parts of completing the collision avoidance function. Whether the system can quickly plan a collision-free path that meets the vehicle dynamics requirements is the key to collision avoidance path planning. The methods commonly used for vehicle emergency active collision avoidance path planning are: planning methods based on curve interpolation description, artificial potential field method, planning methods based on optimization, genetic algorithm, fuzzy and other methods (Chen et al., 2023). Hesse et al. developed a path and velocity distribution map based on potential field and elastic band theory. The method can avoid obstacles in front of the vehicle when reaching the target location (Hesse et al., 2007). However, in real collision avoidance situations, the available data on obstacles is often quite limited. For this reason, some studies have used potential fields to analyze all of the risks in a vehicle's surroundings (Guo et al., 2017; Zhang et al., 2017; Bis et al., 2009; Xia et al., 2021). Kim et al. proposed a method that uses the concept of potential risk. This method identifies the potential risks of the surrounding environment and finds the best path for safety (Kim et al., 2017). Nilsson et al. performed trajectory planning for a high-speed loop scenario with multiple obstacles by means of a convex optimal planning method with transverse-longitudinal decoupling (Nilsson et al., 2015). Rosolia et al. used a trajectory planning method with predictive control of an outer loop nonlinear model (Rosolia et al., 2016). The method generates collision-free trajectories with synthetic inputs based on a simplified vehicle model. The optimization problem is solved by a generalized minimum residual method augmented with a continuation method.

Obstacle avoidance operations in self-driving cars are mainly focused on solving path-planning problems in some regular driving scenarios (Liu et al., 2020; Xia et al., 2022; Zhang et al.,2023; Song et al., 2023). Therefore, their performance may be unsatisfactory in emergency obstacle avoidance situations. P. Lin et al. proposed a new model predictive path planning controller (MPC) combined with PF with a specific trigger analysis algorithm for monitoring traffic emergencies to handle complex traffic scenarios (Lin et al., 2022). This method can ensure safe autonomous driving when dealing with traffic emergencies. The interpolation curve of the discrete optimization-based path planner is too conservative, which will lead to the inability to plan a safe collision avoidance path in emergency situations. Therefore, Yu et al. designed a segmented path planner for emergency obstacle avoidance conditions (Yu et al., 2021). In the first stage of path planning, the smoothest and obstacle avoidance path is obtained using automatic stop path planning. In the second stage, a suitable path for stabilizing the vehicle is found. to determine the safety of vehicle driving based on the inputs of longitudinal acceleration, velocity, and road curvature. Simulation results show that the proposed collision avoidance motion planning method has a safer trajectory at the same longitudinal distance. In order to avoid the problem of emergency scenarios driving on the highway, He et al. established a transverse acceleration model and collision avoidance minimum safety spacing model based on the theoretical analysis of the five polynomial lane change trajectory model and verified the effectiveness and accuracy of steering collision avoidance of self-driving cars on the highway in the simulation platform, which can effectively improve the safety of highway driving (He et al., 2023). In addition to the safety of collision avoidance, the safety of lateral stability is another key issue for self-driving vehicles under high-speed conditions. Hang et al. proposed an integrated path- planning algorithm (Hang et al., 2021). A nonlinear model predictive control is used to optimize the path, and a multivariate Gaussian distribution and polynomial fitting are used to predict the trajectory of moving obstacles. In the algorithm design, a series of constraints are considered, including minimum turning radius, safety distance, control constraints, and tracking error. Simulation results show that the algorithm can handle both static and dynamic collision avoidance as well as lateral stability. And facing active collision avoidance between autonomous vehicles with motion uncertainty and pedestrians, Feng et al. proposed a candidate trajectory planning method considering spatial and temporal sequences, which combines polynomial path planning and speed planning with variable safety speed, based on which safety, stability, and efficiency as well as different driving styles are evaluated from the candidate trajectories optimal trajectory (Feng et al., 2020). The results show that the method is effective in planning safe, stable, and efficient trajectories in emergency situations.

Both emergency steering and emergency braking are crucial for self-driving vehicle collision avoidance systems. It is the basis for the successful introduction of higher levels of autonomous driving and allows the autonomous vehicle to adjust its trajectory planning to its capabilities, external conditions, and knowledge of human errors to improve safety (De Campos et al., 2021; Xia et al., 2021; Zhao et al., 2019). In (Falcone et al., 2007, 2008), it was pointed out that combining the steering system with the braking system can effectively improve the yaw and the lateral stability of an autonomous vehicle. Wang et al. integrated multiple driver assistance function to improve the safety of the system effectively. The effectiveness of the algorithm was verified in simulation experiments (Wang et al., 2022). In summary, the establishment of a perfect vehicle emergency active collision avoidance control algorithm has high practical application prospects as well as research value.

In order to ensure the safety of self-driving vehicles by taking a reasonable collision avoidance approach in high-speed emergencies, this paper proposes a multi- level decision framework collision avoidance algorithm for emergency scenarios. The algorithm overcomes the limitations of braking and steering avoidance in the applicable scenarios, calculates the current vehicle hazard level based on the safety distance model, and adopts combined braking and steering to avoid collisions, which improves the lateral manoeuvrability of the self-driving vehicle. When the vehicle collision risk is high and emergency steering is used, a feasible collision avoidance path is decided by the emergency active collision avoidance path planning module, and finally, the motion control layer completes the whole collision avoidance operation by precise and stable control of the target quantity. The overall architecture of the emergency active collision avoidance control algorithm proposed in this paper is shown in Figure 1. The system is divided into two layers, the upper layer is the decision planning layer, and the lower layer is the motion control layer. The decision planning layer includes two parts, one is the emergency collision avoidance multi-level decision module to adapt to different working conditions, and the other is the emergency active collision avoidance path planning module using the multi-constraint collision avoidance path planning optimization method. In the planning module, this paper proposes a multi-constraint collision avoidance path planning optimization method that considers the drivable area of the vehicle. In this method, the vehicle drivable area considering collision avoidance constraint is finally determined by establishing the prediction formula of kinematic obstacle position and analysing and calculating the relationship between collision hazard moment and position constraint. Then, the optimization function considering the kinematic constraints as well as the position constraints is established, and the final collision avoidance path is determined through the optimization solution. The lower motion control layer establishes the vehicle transverse and longitudinal motion control module, including vehicle longitudinal acceleration control based on feedforward plus feedback algorithm and MPC path tracking control based on the vehicle dynamics model. In addition, the raw information acquired by the sensors is processed so that the algorithm obtains stable and effective sensory information during the execution.

**Figure 1** Emergency active collision avoidance control algorithm architecture.

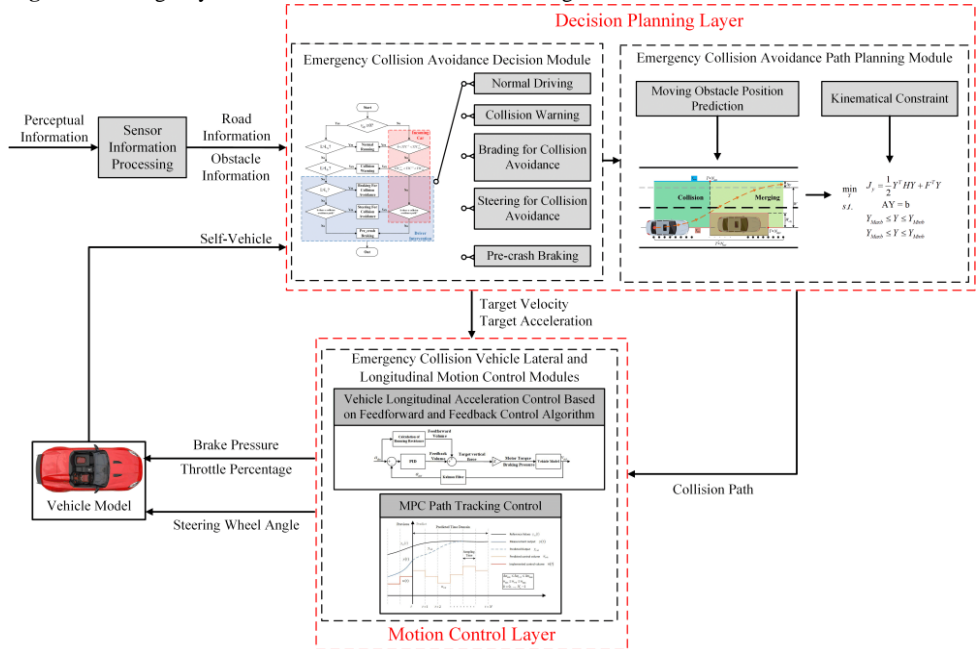

The remainder of this paper is organized as follows: In section 2 the designed multi-level obstacle avoidance decision algorithm is proposed. In section 3, the multi-constraint collision avoidance path planning optimization method proposed in the decision algorithm is provided. In section 4, respectively provides the results of simulation and experimental studies. Finally, section 5 summarizes the contributions of the paper and puts forward suggestions for future work. Our contributions are as follows:

(1) For improved lateral manoeuvrability, this paper proposes a multi-level decision avoidance framework that integrates steering and braking systems. The system takes the optimal operation according to different risk levels to avoid the high-risk situation of vehicle instability due to the separation of steering and braking actions.

(2) In the planning module of steering obstacle avoidance operation, this paper proposes a multi-constraint collision avoidance path planning optimization method considering the drivable area of the vehicle, defines the collision hazard moment, and analyses and calculates the relationship between the collision hazard moment and the position constraint, finally establishes the drivable area that meets the actual collision avoidance requirements and obtains the planning path. The method reduces the complexity of the algorithm, realizes the precise restriction of the lane change process, satisfies the requirements of the obstacle avoidance framework for computational efficiency, and takes into account the driver's comfort.

(3) In order to verify the effectiveness of the algorithm, this paper systematically composes four typical hazardous traffic scenarios under structured roads: collision avoidance by stationary obstacles, collision avoidance by the front car emergency stop, collision avoidance by pedestrians crossing lanes and collision avoidance by opposite direction traffic. The system can accomplish the collision avoidance task under different scenarios, and the algorithm has high flexibility and scenario-using capability.

## 2    Design of Multilevel Collision Avoidance Decision Algorithm

The task of the vehicle emergency active collision avoidance decision algorithm is to make behavioral decisions for vehicles in hazardous situations by using the current vehicle state, vehicle surroundings, obstacle information, and driver's behavior. For different dangerous traffic scenarios and collision risk levels, the vehicle can take collision avoidance operations according to the set collision avoidance decision logic to improve safety.

**Figure 2** General framework of multilevel collision avoidance decision algorithm.

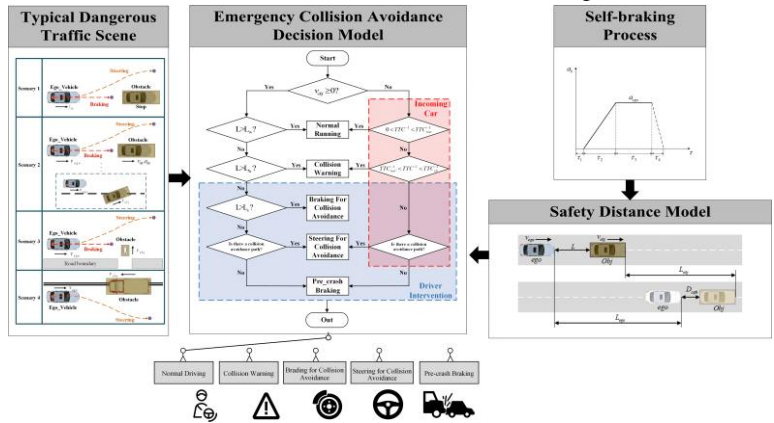

**Figure 3** Vehicle collision avoidance hazard classification diagram.

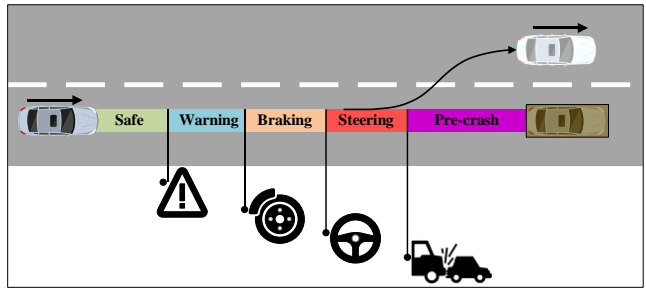

The emergency active collision avoidance decision algorithm first analyses the vehicle braking process through the motion formula. At the same time, the target braking end state of the self-vehicle is divided into two categories: self-vehicle braking speed is zero and self-vehicle braking speed is the target speed. The corresponding vehicle braking safety distance formula is established, as shown in Figure 3. Based on the braking distance formula and the definition of the minimum safe distance of the vehicle. This paper establishes the distance-switching threshold between the four collision avoidance behaviors of the vehicle. Finally, a collision avoidance decision algorithm that considers typical hazard scenarios and satisfies the driver intervention exit mechanism is established. The decision algorithm carries out the corresponding collision avoidance operation by calculating the current hazard level of the self-vehicle. When the path planning module is unable to calculate an effective collision avoidance path, the system selects the maximum braking force for pre-crash braking to minimize the damage caused by the collision.

## 2.1 Collision avoidance safety distance model

As shown in Figure 4, when a vehicle takes an emergency braking operation, the change in acceleration generally consists of four phases.

**Figure 4** Vehicle braking acceleration change process.

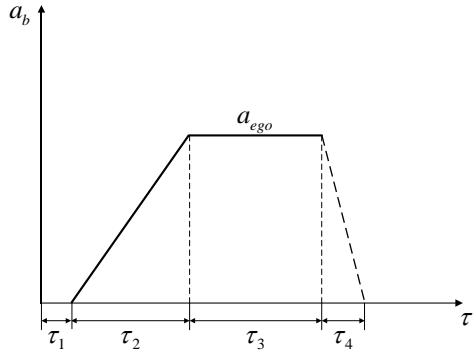

The vehicle braking process is modeled as follows:

$$D_b = \begin{cases} \frac{1}{3.6}\left(\tau_1 + \frac{\tau_2}{2}\right)v_{ego} + \frac{v_{ego}^2}{25.92 a_{ego}} & v_{ego\_end} = 0 \\ \frac{1}{3.6}\left(\tau_1 + \frac{\tau_2}{2}\right)v_{ego} + \frac{v_{ego}^2 - v_{ego\_end}^2}{25.92 a_{ego}} & v_{ego\_end} \neq 0 \end{cases} \quad (1)$$

where $v_{ego}$ is the self-vehicle speed, $a_{ego}$ is the self-vehicle target braking deceleration, $v_{ego\_end}$ is the target vehicle speed at the end moment, $\tau_1$ is the time of braking system adjustment, and $\tau_2$ is the time of braking deceleration growth.

**Figure 5** Vehicle minimum safety distance diagram.

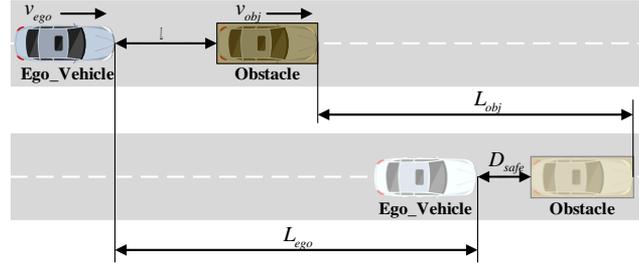

Figure 5 for the vehicle minimum safety distance diagram. $D_{safe}$ is the end of the vehicle's active collision avoidance, the relative distance between the vehicle and the obstacles in front, $L_{ego}$ is the braking distance of the vehicle, $L_{obj}$ is the distance of the obstacles in front of travel. Then the vehicle safety distance L can be expressed as follows：

$$L = L_{ego} - L_{obj} + D_{safe} \tag{2}$$

The vehicle safety distance model established in this paper is mainly used to distinguish four vehicle target operations, namely safe driving, forward collision warning, emergency braking, and emergency steering. Then determine the safety distance of warning, the safety distance of starting braking, and the minimum safety distance of braking for the four cases. Assuming that the front obstacle movement state is kept constant, combined with the analysis of the vehicle braking process, the obstacle movement process is divided into three cases: stationary, uniform motion or accelerated motion, and emergency braking. For the three cases of safety distance calculation, the derivation of the calculated safety distance model can be adapted to the vehicle in front of the obstacles in different states of motion, as shown in Table 1.

**Table 1** Minimum safety distance of collision avoidance for different obstacle motion states

| Status of obstacles ahead | $\begin{cases} \text{the safety distance of warning } L_w \\ \text{the safety distance of starting braking } L_b \\ \text{the minimum safety distance of braking } L_s \end{cases}$ |
|---|---|
| 1) Stationary | $\begin{cases} \frac{1}{3.6}\left(\tau_1 + \frac{\tau_2}{2}\right)v_{ego} + \frac{v_{ego}^2}{25.92 a_{ego\_min}} + D_{safe} + \frac{1}{3.6}t_{driver}v_{ego} \\ \frac{1}{3.6}\left(\tau_1 + \frac{\tau_2}{2}\right)v_{ego} + \frac{v_{ego}^2}{25.92 a_{ego\_min}} + D_{safe} \\ \frac{1}{3.6}\left(\tau_1 + \frac{\tau_2}{2}\right)v_{ego} + \frac{v_{ego}^2}{25.92 a_{ego\_max}} + D_{safe} \end{cases}$ |
| 2) Uniform motion or accelerated motion | $\begin{cases} \frac{1}{3.6}\left(\tau_1 + \frac{\tau_2}{2}\right)(v_{ego} - v_{obj}) + \frac{v_{ego}^2 - v_{obj}^2}{25.92 a_{ego\_min}} + D_{safe} + \frac{1}{3.6}t_{driver}v_{ego} \\ \frac{1}{3.6}\left(\tau_1 + \frac{\tau_2}{2}\right)(v_{ego} - v_{obj}) + \frac{v_{ego}^2 - v_{obj}^2}{25.92 a_{ego\_min}} + D_{safe} \\ \frac{1}{3.6}\left(\tau_1 + \frac{\tau_2}{2}\right)(v_{ego} - v_{obj}) + \frac{v_{ego}^2 - v_{obj}^2}{25.92 a_{ego\_max}} + D_{safe} \end{cases}$ |
| 3) Emergency braking | $\begin{cases} \frac{1}{3.6}\tau_1 v_{ego} + \frac{1}{7.2}\tau_2(v_{ego} - v_{obj}) + \frac{v_{ego}^2}{25.92 a_{ego\_min}} - \frac{v_{obj}^2}{25.92 a_{obj}} + D_{safe} + \frac{1}{3.6}t_{driver}v_{ego} \\ \frac{1}{3.6}\tau_1 v_{ego} + \frac{1}{7.2}\tau_2(v_{ego} - v_{obj}) + \frac{v_{ego}^2}{25.92 a_{ego\_min}} - \frac{v_{obj}^2}{25.92 a_{obj}} + D_{safe} \\ \frac{1}{3.6}\tau_1 v_{ego} + \frac{1}{7.2}\tau_2(v_{ego} - v_{obj}) + \frac{v_{ego}^2}{25.92 a_{ego\_max}} - \frac{v_{obj}^2}{25.92 a_{obj}} + D_{safe} \end{cases}$ |

where the relative distance at the end of the braking moment $D_{safe}$ is:

$$D_{safe} = \begin{cases} 3.6 & v_{ego} = 0 \\ max(0.2364 v_{ego} + 1.6109, 3.6) & v_{ego} > 0 \end{cases} \quad (3)$$

The trigger braking deceleration $a_{ego\_min}$ and the maximum braking deceleration $a_{ego\_max}$ are:

$$\begin{cases} a_{ego\_min} = min(4m/s^2, \mu g) \\ a_{ego\_max} = max(7m/s^2, \mu g) \end{cases} \quad (4)$$

*2.2 Multilevel collision avoidance decision logic based on crash risk assessment*

For the side of the opposite lane vehicles across the centreline approaching conditions, emergency braking is not effective to avoid the collision. At this time, only emergency steering can be taken to avoid the collision. However, when encountering a situation where the speed of the oncoming vehicle is too low or the relative distance is small, steering to avoid collision is not in line with the general driver's operating habits. Therefore, in this paper, we establish the method of TTC risk assessment. The inverse of the collision time is used to characterize the degree of collision risk when a vehicle in the opposite lane crosses the centerline and approaches, which is calculated as follows:

$$TTC^{-1} = \frac{v_{ego} - v_{obj}}{L} \quad (5)$$

where $v_{ego}$ is the self-vehicle speed, $v_{obj}$ is the speed of the obstacle vehicle, and $L$ is the relative distance between the vehicle and the proximal face of the obstacle. In this paper, based on Sun Y's classification of the inverse of the collision time corresponding to the collision risk level, the collision warning threshold $TTC_{war}^{-1}$ is set to 0.3, and the active steering collision avoidance threshold $TTC_{str}^{-1}$ is 0.5.

**Figure 6** Emergency active collision avoidance decision logic diagram.

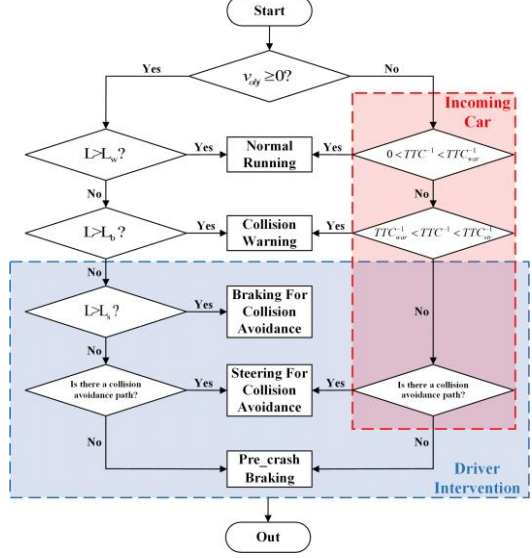

For the oncoming vehicle working condition, when $TTC_{war}^{-1} < TTC^{-1} < TTC_{str}^{-1}$, the vehicle is considered to be at risk of collision with the vehicle coming from the opposite direction, but the risk is small, at this time the system only carries out a collision warning to the driver. When $TTC^{-1} > TTC_{str}^{-1}$, emergency steering is taken to avoid the collision. And when the risk of vehicle collision is too large, the current environment does not exist an effective steering collision avoidance path, the system believes that there is no longer collision avoidance, then the system with the maximum braking force braking for pre-collision. The goal is to minimize the damage caused by the collision. Outside the entire system, the driver's right to take over is ensured. The system automatically exits vehicle control when the system detects driver intervention. The decision logic of the vehicle emergency active collision avoidance control algorithm is shown in Figure.6, where $L$ is the relative distance and $v_{obj}$ is the speed of the obstacle vehicle.

## 3   Evasive maneuver planning with safety envelops

In the vehicle collision avoidance process path planning algorithm, the safety constraints include the calculation of the drivable area and dynamics constraints. The path in this chapter is solved by the planning algorithm optimization function to obtain a collision avoidance path that meets the requirements of collision avoidance.

*3.1 Description of vehicle drivable area*

As shown in Figure 7, during the vehicle emergency active collision avoidance, the vehicle drivable area is determined by the lateral position of the obstacle $W_{obj}$ and the lane change width $W$. The self-vehicle position needs to be between the upper boundary $Y_{max}$ and the lower boundary $Y_{min}$. These upper and lower boundaries are influenced by the movement state of the obstacle. For stationary obstacles, the vehicle travelable area is fixed; for moving obstacles, the vehicle travelable area needs to be combined with the prediction of the obstacle position and the correction of the original travelable area constraint.

**Figure 7** Diagram of vehicle drivable area.

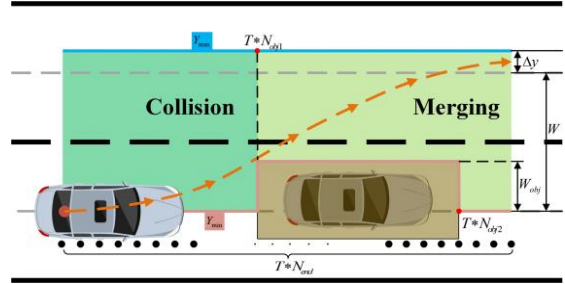

According to the relative position of the self-vehicle and the obstacle in the collision avoidance process, the collision avoidance process is divided into two phases: the collision avoidance phase and the merging phase. These two phases demarcation point for the collision hazard moment $T * N_{obj1}$. After this moment, the vehicle's lateral position crosses the left end face of the obstacle, the vehicle is considered to have no collision threat. When the vehicle is in the collision avoidance phase, the main purpose of the system at this time is to avoid a collision between the self-vehicle and the obstacle. When the vehicle crosses the obstacle, the vehicle enters the merging stage. At this time the vehicle has no collision threat, where $T * N_{obj2}$ for the vehicle's longitudinal position over the front surface of the obstacle moment. Since in the merging phase, the size of the longitudinal distance is related to the drivable area detected by the sensor. Therefore, the minimum value of the longitudinal distance should be greater than the longitudinal position of the far-end face of the obstacle. However, the distance is too large to meet the driving habits of the driver when merging, so in the merging phase, the longitudinal distance should be greater than or equal to the longitudinal distance in the collision avoidance phase.

*3.1.1 Determination of the drivable area of stationary obstacles*

The position constraint established based on the drivable area of the vehicle can be expressed as:

$$Y_{max} = W \quad \forall t \in 1, \cdots, N_{end}$$
$$Y_{min} = \begin{cases} 0 & \forall t \in 1, \ldots, N_{obj1} \\ W_{obj} & \forall t \in N_{obj1}, \ldots, N_{obj2} \\ 0 & \forall t \in N_{obj2}, \ldots, N_{end} - 1 \\ W & t = N_{end} \end{cases} \quad (6)$$

In the merging phase, the vehicle slowly moves into the target lane and gradually enters the lane-keeping state. The collision risk at this time is low, so there is no strict requirement for the lateral position. On the premise of not affecting the collision avoidance effect, to reduce the computational burden of the planning algorithm, $\Delta y$ is set as the allowed lateral

(7)

deviation of the vehicle at the end of the lane change moment. The vehicle is allowed to have a certain range of deviation from the target lane at that moment. Therefore, the position constraint of the vehicle can be expressed as :

$$Y_{max} = W + \Delta y \quad \forall t \in 1,\ldots,N_{end}$$

$$Y_{min=} \begin{cases} 0 & \forall t \in 1,\ldots,N_{obj1} \\ W_{obj} & \forall t \in N_{obj1},\ldots,N_{obj2} \\ 0 & \forall t \in N_{obj2},\ldots,N_{end} - 1 \\ W - \Delta y & t = N_{end} \end{cases}$$

When establishing the location constraints for path planning, the self-vehicle and the obstacle cannot be considered as mass points, and the influence of the size of the self-vehicle and the obstacle on the constraints needs to be considered. This is to ensure that the self-vehicle does not collide with the obstacle when it travels along the planned path. Therefore, this paper defines the collision hazard moment in the collision avoidance process, as shown in Figure 8.

**Figure 8** Diagram of the collision hazard moment.

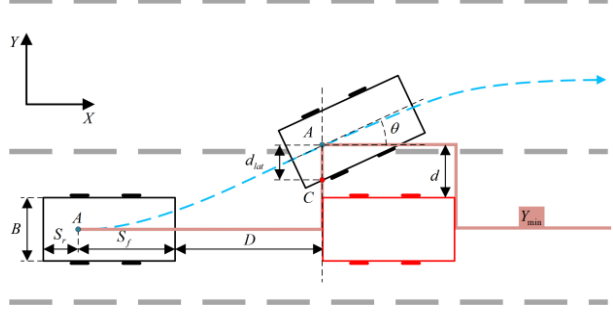

In Figure 8, point $A$ is the midpoint of the rear axle of the vehicle, point $C$ is the edge of the vehicle in the longitudinal position of the midpoint of the rear axle, and the nearest point with the obstacle, in this paper point B is called the collision risk point, $\theta$ is the heading angle of the vehicle, $S_f$, $S_r$ are the distance from point A to the front and rear end of the vehicle, $D$ is the distance between the vehicle and the near end of the obstacle, $d$ is the minimum lateral distance between the midpoint of the rear axle of the vehicle and the obstacle set to avoid the collision.

The collision hazard moment is defined as the moment when the longitudinal position of the midpoint of the rear axle of the self-vehicle coincides with the position of the proximal surface of the obstacle. Because of the collision avoidance process, the longitudinal speed of the vehicle is much larger than the lateral speed, so in the vehicle over the obstacle car this time, the vehicle's heading angle change is small. And in a short period, the lateral relative distance between the self-car and the obstacle changes less, so do not consider the case of the obstacle hitting the side of the self-vehicle. As the vehicle heading angle increases, the distance of point $C$ from the obstacle is becoming smaller. Therefore, setting a reasonable $d$ value, can ensure that the vehicle in the maximum heading angle $C$ point does not intersect with the obstacle, can avoid the collision, and can be expressed in the following formula：

$$d > \frac{B}{2 \cos \theta_{max}} \tag{8}$$

*Multi-level decision framework collision avoidance algorithm in emergency scenarios* 13

where the maximum heading angle $\theta_{max}$ is calculated from the angle between the set maximum lateral velocity and longitudinal velocity:

$$\theta_{max} = arctan \frac{v_{ymax}}{v_x} \quad (9)$$

In the collision avoidance process, if the lateral position of the collision danger moment crosses the left end face of the obstacle, it can be considered that no collision occurs with the obstacle. Based on the collision danger moment to establish the vehicle can drive area position constraint since the vehicle and the relative position information of the obstacle can be effectively transformed into position constraint to ensure the safety of the collision avoidance process.

### 3.1.2 Drivable area correction for moving obstacle conditions based on collision hazard moments

In the emergency active collision avoidance process, the position of the moving obstacle changes with time. This situation leads to the obstacle position constraint does not meet the actual collision avoidance requirements. Therefore, in the emergency collision avoidance path planning process, this paper converts the position of the moving obstacles at the moment of the actual collision risk into the corresponding position constraint by determining the actual collision risk moment, and then the initial position constraint is modified. In different obstacle motion cases, the position constraint process is established with the collision risk moment, as shown in Figure 9-Figure 11.

**Figure 9** Schematic diagram of the position constraint of the opposing motion of the obstacle.

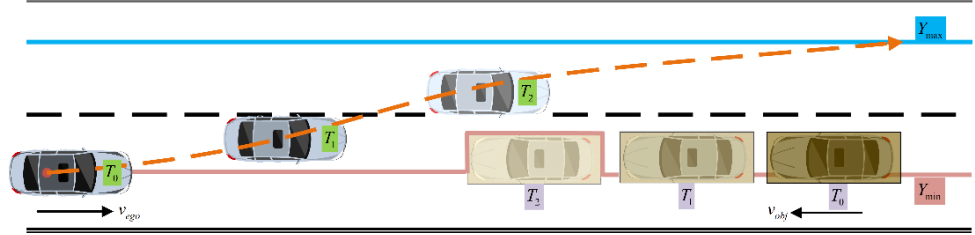

**Figure 10** Schematic diagram of the Obstacle isotropic motion position constraint.

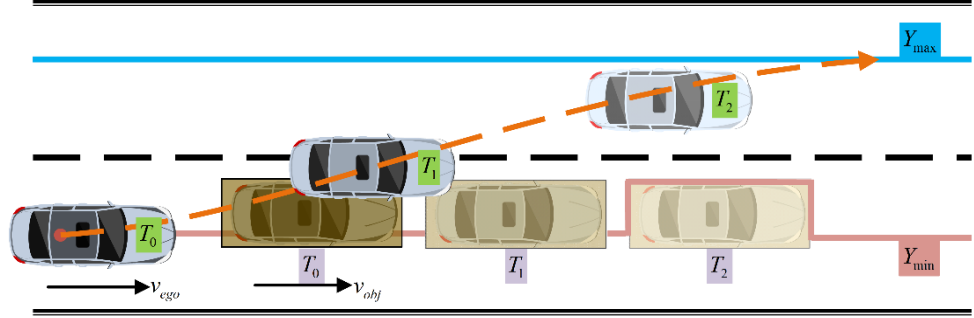

**Figure 11** Obstacle lateral drive-in position constraint schematic.

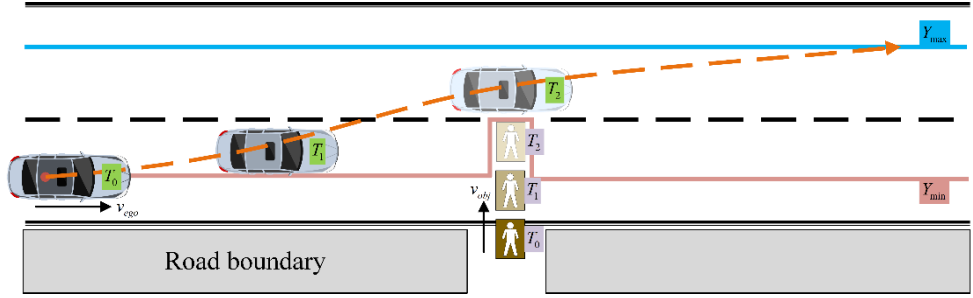

According to the above front obstacle movement scenario, the calculation of the obstacle collision hazard moment needs to consider the relative motion state of the self-car and the obstacle along the lane direction. The results of the calculation are shown in the following Table 2:

**Table 2** Motion obstacle collision hazard moment calculation

| Longitudinal acceleration of the obstacle | Collision hazard moment | |
|---|---|---|
| $a_{obj} \geq 0$ | $T_2 = \dfrac{L}{v_{ego} - v_{obj}}$ | |
| $a_{obj} < 0$ | $v_{ego} t_{brake} > L + x_{brake}$ | $v_{ego} t_{brake} \leq L + x_{brake}$ |
| | $T_2 = \dfrac{v_{ego}^2 + 2aL}{2av_{ego}}$ | $T_2 = \dfrac{v_{obj} - v_{ego} + \sqrt{(v_{obj} - v_{ego})^2 + 2a_{obj}L}}{a}$ |

To ensure the safety as well as the comfort of vehicle driving in the process of emergency active collision avoidance, this paper sets up the cost functions of vehicle lateral speed, lateral acceleration, and lateral jerk:

$$J_y = \sum_{t=1}^{N_{end}}(p_t v_{y_t}^2 + q_t a_{y_t}^2 + r_t j_{y_t}^2) \qquad (10)$$

where $p_t$, $q_t$, and $r_t$ are the weight values corresponding to the lateral velocity, lateral acceleration, and lateral jerk of the collision avoidance path, respectively. The smoothness and feasibility of the collision avoidance path are ensured by setting reasonable non-negative weight values.

$$\begin{cases} y_{min_t} \leq y_t \leq y_{max_t} & \forall t \in 1, \cdots, N_{end} \\ v_{min_t} \leq v_{y_t} \leq v_{max_t} & \forall t \in 1, \cdots, N_{end} \\ a_{min_t} \leq a_{y_t} \leq a_{max_t} & \forall t \in 1, \cdots, N_{end} \\ j_{min_t} \leq j_{y_t} \leq j_{max_t} & \forall t \in 1, \cdots, N_{end} \end{cases} \qquad (11)$$

where $y_t, v_{y_t}, a_{y_t}, j_{y_t}$ represent the lateral position, lateral velocity, lateral acceleration, and lateral jerk of the self-vehicle at the sampling moment $t$. $y_t$ needs to satisfy the vehicle position constraint and requires each moment to ensure that the path lateral position is within the vehicle travelable area. $v_{y_t}, a_{y_t}, j_{y_t}$ all need to satisfy the kinematic constraints of the vehicle and are influenced by the vehicle travel constraints and the road adhesion coefficient.

In the emergency active collision avoidance process, $\varphi_y(t)$ is used to represent the lateral motion state of the vehicle at sampling moment $t$, and $Y$ denotes the set of vehicle motion states at all sampling moments in the lane change process.

$$\varphi_y(t) = [y_t, v_{y_t}, a_{y_t}, j_{y_t}]^T \quad \forall t \in 1, \ldots, N_{end} \tag{12}$$

$$Y = [\varphi_y(1), \ldots, \varphi_y(N_{end})]^T \tag{13}$$

where $b_{max}, b_{min}$ denote the constraints on the lateral motion state of the vehicle at the $t$ sampling moments, and with $B_{max}, B_{min}$ denote the boundaries of the vehicle motion state at all sampling moments during the lane change.

$$\begin{cases} b_{max}(t) = [y_{max}, v_{max}, a_{max_t}, j_{max_t}]^T & \forall t \in 1, \cdots, N_{end} \\ b_{mIN}(t) = [y_{min_t}, v_{min_t}, a_{min_t}, j_{min_t}]^T & \forall t \in 1, \cdots, N_{end} \end{cases} \tag{14}$$

$$\begin{cases} B_{max} = [b_{max}(1), \cdots, b_{max}(N_{end})]^T \\ B_{min} = [b_{min}(1), \cdots, b_{min}(N_{end})]^T \end{cases} \tag{15}$$

From the above equation, the motion state constraint of the vehicle during the whole lane change can be expressed by the following equation:

$$B_{min} \leq Y \leq B_{max} \tag{16}$$

During the lane change process, each moment of the vehicle path is required to satisfy the physical motion equation. The lateral displacement at the current moment is obtained by accumulating the lateral displacements in each previous sampling step, and the relationship between each sampling step of the vehicle is expressed as follows:

$$\begin{cases} y_{t+1} = y_t + v_{y_t}T + a_{y_t}\frac{T^2}{2} & \forall t \in 1, \ldots, N_{end} - 1 \\ v_{y_{t+1}} = v_{y_t} + a_{y_k}T & \forall t \in 1, \ldots, N_{end} - 1 \\ a_{y_{t+1}} = a_{y_t} + j_{y_t} & \forall t \in 1, \ldots, N_{end} - 1 \end{cases} \tag{17}$$

Simplify it to:

$$a_{nb} \begin{bmatrix} \varphi_y(t) \\ \varphi_y(t+1) \end{bmatrix} = b \quad \forall k \in 1, \ldots, N_{end-1} \tag{18}$$

Among them:

$$a_{nb} = \begin{bmatrix} 1 & T & \frac{T^2}{2} & 0 & -1 & 0 & 0 & 0 \\ 0 & 1 & T & 0 & 0 & -1 & 0 & 0 \\ 0 & 0 & 1 & 1 & 0 & 0 & -1 & 0 \end{bmatrix} \quad b = [0 \ \ 0 \ \ 0] \tag{19}$$

The motion state constraint of the vehicle between each sampling step is represented as follows：

$$AY = b \tag{20}$$

Among them:

$$A = \begin{bmatrix} a_{nb} & 0 & 0 & 0 \\ 0 & a_{nb} & 0 & 0 \\ 0 & 0 & \ddots & \vdots \\ 0 & 0 & \cdots & a_{nb} \end{bmatrix}_{3(N_{end}-1) \times (4 \times N_{end})} \tag{21}$$

$$b = [0 \ \ 0 \ \cdots \ 0]^T_{3(N_{end}-1)}$$

In summary, the standard form of secondary planning can be obtained：

$$\min_Y J_y = \frac{1}{2}Y^T HY + F^T Y \tag{22}$$

$$s.t. \quad \begin{matrix} AY = b \\ B_{min} \leq Y \leq B_{max} \end{matrix}$$

Among them:

$$h_t = \begin{bmatrix} 0 & 0 & 0 & 0 \\ 0 & 2p_t & 0 & 0 \\ 0 & 0 & 2q_t & 0 \\ 0 & 0 & 0 & 2r_t \end{bmatrix} \quad H = \begin{bmatrix} h_1 & 0 & 0 & 0 \\ 0 & h_2 & 0 & 0 \\ 0 & 0 & \ddots & \vdots \\ 0 & 0 & \cdots & h_{end} \end{bmatrix}$$

$$f_t = \begin{bmatrix} 0 \\ 0 \\ 0 \\ 0 \end{bmatrix} \quad F = \begin{bmatrix} f_1 \\ f_2 \\ \vdots \\ f_{end} \end{bmatrix}$$

(23)

The path planning problem of the vehicle emergency active collision avoidance process is converted into the standard form of quadratic programming described in the above equation, and by setting reasonable weight parameters and using optimization methods such as the interior point method or the effective set method to solve, a collision avoidance path that meets the collision avoidance requirements can be obtained.

## 4  Simulation verification

To study the performance of the proposed vehicle emergency active collision avoidance control algorithm, this section first summarizes typical vehicle obstacle avoidance scenarios, as shown in Table 3, and introduces the simulation results of the algorithm in different scenarios.

In Table 3, the relative distances of high and low collision risk scenarios and the lateral distance of obstacles are the relative distances between the self-driving car and the obstacles when the system determines the presence of collision risk. Since the parameters of the stationary obstacle collision avoidance condition and the front car emergency stop collision avoidance condition are similar, and the latter is more representative, due to the limitation of space, this section selects a typical representative of the obstacle avoidance condition for simulation experiments.

**Table 3** Emergency steering test scenario parameters

*Multi-level decision framework collision avoidance algorithm in emergency scenarios 17*

| Collision Avoidance Scenarios | Front Car Emergency Brake | Pedestrians Crossing Lanes | Opposite Direction Traffic | Unit |
|---|---|---|---|---|
| Self-vehicle Speed | 25 | 22.2 | 16.7 | m/s |
| Obstacle Speed | 16.7 | 1.4 | 16.7 | m/s |
| Obstacle Acceleration | -7 | 0 | 0 | m/s$^2$ |
| Direction of Obstacle Movement | In the same direction | To the left | Reverse direction | -- |
| Obstacle Lateral Distance | 0 | 2 | 1 | m |
| Obstacle Size (length × width) | 4.5×1.9 | 0.4×0.6 | 4.5×1.9 | m |
| Collision Avoidance Longitudinal Distance | 120 | 120 | 90 | m |
| Trigger Braking Safety Distance | 75.7 | 67.6 | -- | m |
| Minimum Safety Distance for Braking | 42.3 | 41.1 | -- | m |
| Relative Distance of High-risk Scenarios | 26 | 30 | -- | m |
| Relative Distance of Low-risk Scenarios | 60 | 55 | 100 | m |

*4.1 Scenario I: Front car emergency brake collision*

*4.1.1 Simulation verification of high collision risk conditions for front vehicle emergency brake*

The simulation results of the front vehicle emergency brake's high collision risk condition are shown in Figure 12.

**Figure 12** The process of collision avoidance for high collision risk working conditions in the front car emergency brake.

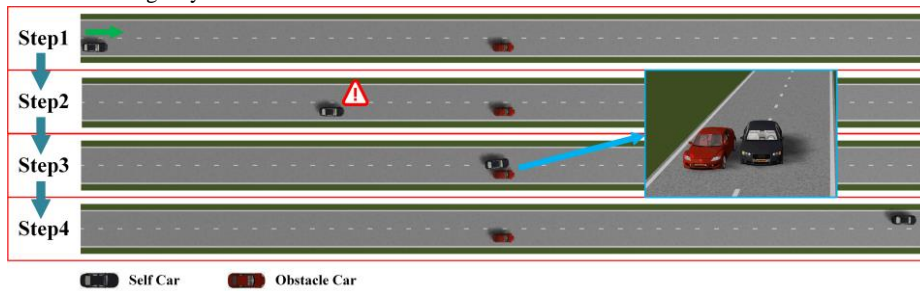

In the Step1 stage, the vehicle is driving normally at 25 m/s speed, and the obstacle vehicle in front is driving at a uniform speed of 16.7 m/s. In the Step2 stage, the obstacle vehicle suddenly brakes urgently with a reduced speed of 7 m/s$^2$. At this time, the longitudinal relative distance between the vehicle and the obstacle vehicle is 26 m, while the minimum safe distance of braking is 42.3 m, and the risk of collision is large. The collision cannot be avoided by emergency braking, and the system carries out collision avoidance operation by emergency steering, and the system plans the collision avoidance path and carries out tracking. In the Step3 stage, the vehicle successfully crosses the obstacle vehicle and continues the merging process. In the Step4 stage, the vehicle completes the collision avoidance process.

The results of collision avoidance path planning are shown in Figure 13:

**Figure 13** Planning path and constraint settings for high collision risk conditions of front vehicle emergency brake.

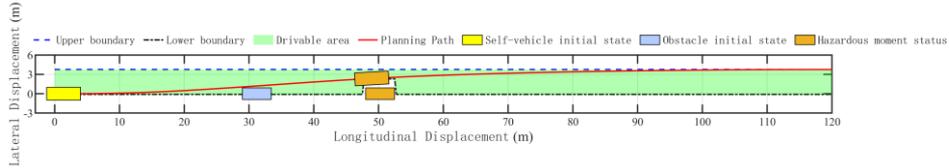

The results of vehicle tracking for the collision avoidance path are shown in Figure14:

**Figure 14** Path tracking results of high collision risk conditions of the front vehicle emergency brake.

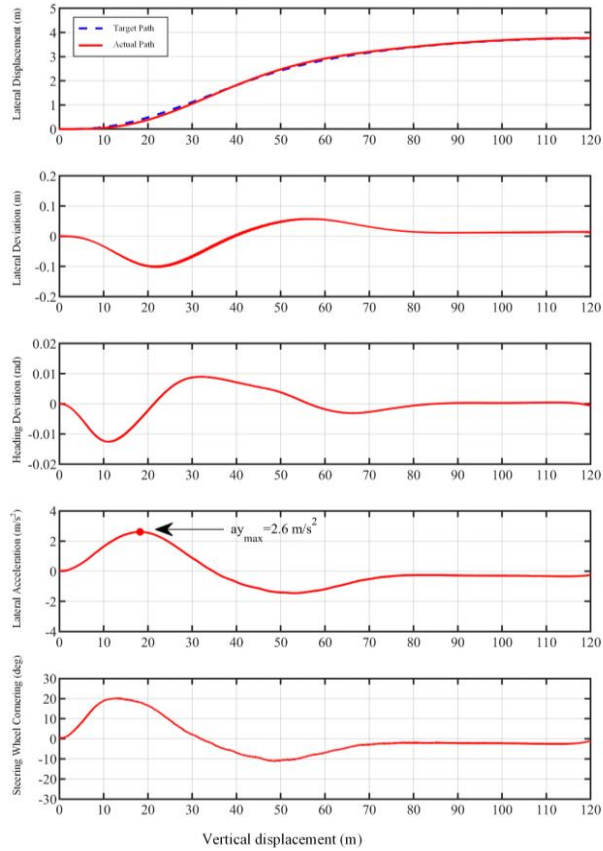

From the tracking results, it can be concluded that the maximum lateral acceleration is 2.6 m/s$^2$, the lateral deviation is less than 0.1 m, the heading deviation is kept within 0.01 rad, and the steering wheel angle changes smoothly.

*4.1.2  Simulation verification of low-crash risk conditions for front vehicle emergency brake*

The simulation result curve of the front vehicle emergency brake's low collision risk condition is shown in Figure 15. From the simulation results can be obtained, the vehicle to 25 m/s speed uniform speed, and the speed of the obstacle vehicle is 16.7 m/s. At this time, the vehicle and the longitudinal relative distance of the obstacle vehicle ahead is greater than the trigger braking safety distance under this condition, the vehicle for normal driving. When the front obstacle vehicle suddenly emergency braking with 7 m/s$^2$ deceleration speed, at this time the emergency collision avoidance decision algorithm gets the trigger braking safety distance and braking minimum safety distance jump, respectively to 75.7 m and 42.3 m. And then the relative distance is 60 m, and the vehicle into an emergency braking state, with a 4 m/s$^2$ target deceleration speed for braking. When the actual distance is equal to the minimum safety distance of braking, the vehicle brakes with the maximum braking deceleration speed of 7 m/s$^2$ until the vehicle stops. The longitudinal relative distance between the end moment of collision avoidance and the front obstacle vehicle is 4.1 m, effectively avoiding the collision accident.

**Figure 15**  Collision avoidance results of the front vehicle emergency stop with low collision risk conditions.

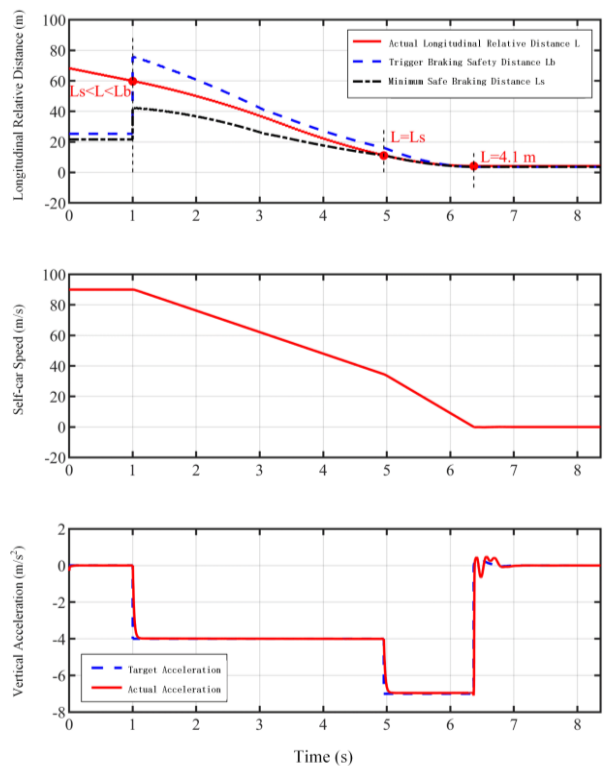

*4.2 Pedestrian crossing lane collision avoidance conditions*

*4.2.1 Simulation Validation of High Crash Risk Conditions for Pedestrians Crossing Lanes*

The simulation results of the pedestrian crossing lane collision avoidance condition are shown in Figure 16. In the Step1 stage, the vehicle initially drives normally at 22.2 m/s, and the pedestrian moves up along the vertical direction of the lane at 1.4 m/s on the right side of the house. In Step 2, the pedestrian crosses the house to the side of the lane. At this time, the self-vehicle detects pedestrian information. The longitudinal relative distance between the vehicle and the pedestrian is 29.7 m. The lateral relative distance is -2 m. And the minimum safety distance for braking is 40.8 m. Due to the high risk of collision, the system adopts emergency steering for collision avoidance. The vehicle plans the collision avoidance path and tracks it. In the Step3 stage, the vehicle successfully crosses the pedestrian and continues the vehicle merging. In Step 4, the vehicle completes the collision avoidance process. The results of collision avoidance path planning are shown in Figure 17.

**Figure 16** Collision avoidance process for pedestrian crossing with high crash risk conditions.

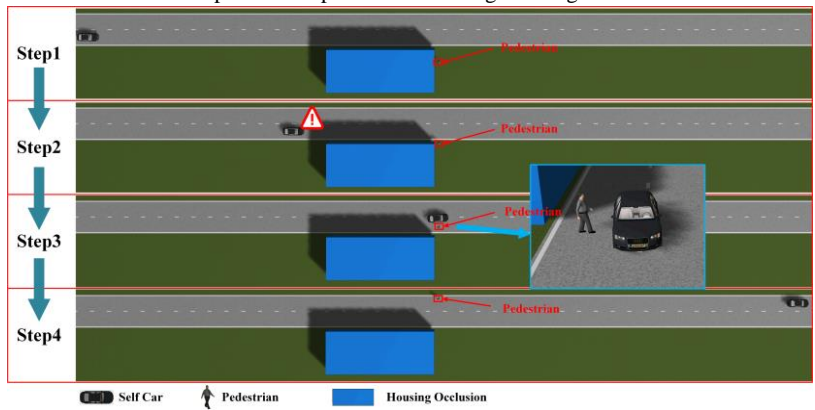

**Figure 17** Planning path and constraint settings for high collision risk conditions for pedestrian crossing lanes.

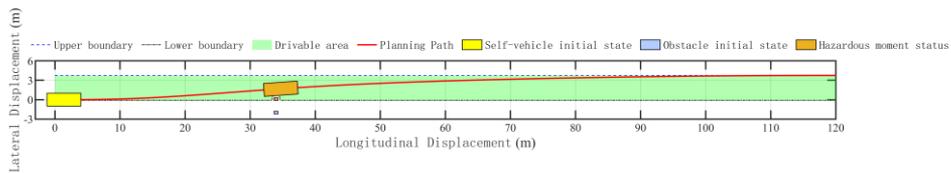

The tracking results of the vehicle for the collision avoidance path are shown in the following figure as shown in Figure 18. From the tracking results, we can see that the maximum lateral acceleration of tracking is 2.5 m/s$^2$, the lateral deviation of tracking is less than 0.1 m, and the deviation of heading is kept within 0.015 rad. The steering wheel angle changes smoothly.

**Figure 18** Path tracking results for high collision risk conditions for pedestrians crossing lanes.

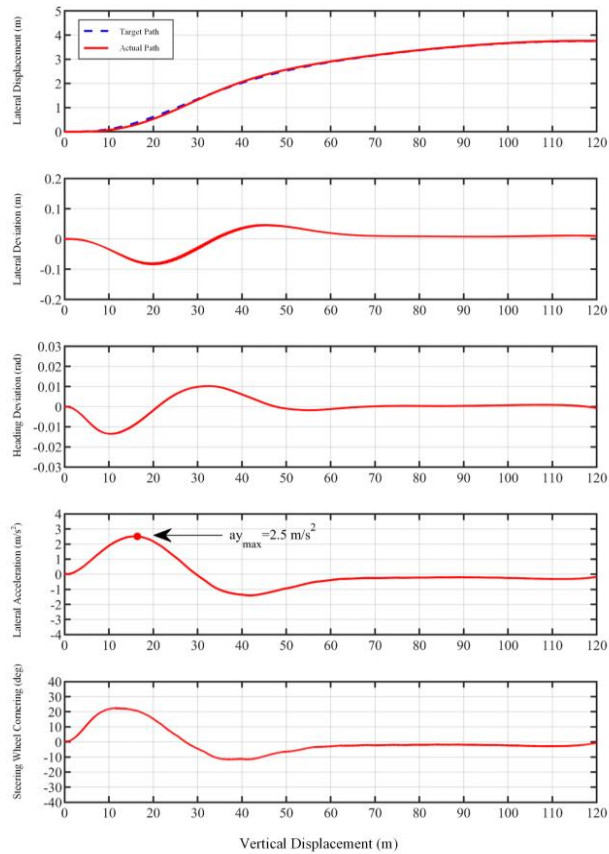

*4.2.2 Simulation Validation of Low Crash Risk Conditions for Pedestrians Crossing Lanes*

The simulation result curve for the low collision risk condition of the pedestrian crossing lane is shown in Figure 19. From the simulation results in the figure, it can be concluded that no pedestrian is detected by the vehicle at the beginning moment. The vehicle is driving at a constant speed of 22.2 m/s. When the pedestrian is detected, the longitudinal relative distance between the vehicle and the pedestrian is 55 m. This distance is less than the current trigger braking safety distance (67.6 m) and greater than the current braking minimum safety distance (41.1 m), and the vehicle enters the emergency braking state at this time. The vehicle starts braking at a target deceleration speed of 4 m/s$^2$. With the change of the self-vehicle speed and the longitudinal relative distance with the pedestrian, the trigger braking safety distance and the braking minimum safety distance calculated by the emergency collision avoidance decision algorithm also change. When the actual distance is equal to the braking minimum safety distance, the vehicle brakes at the set maximum braking deceleration rate of 7 m/s$^2$ until the vehicle stops completely. And the relative longitudinal distance between the vehicle and the pedestrian at the end of braking is greater than 0, which effectively avoids the collision accident.

**Figure 19** Collision avoidance results for low crash risk conditions for pedestrian crossings.

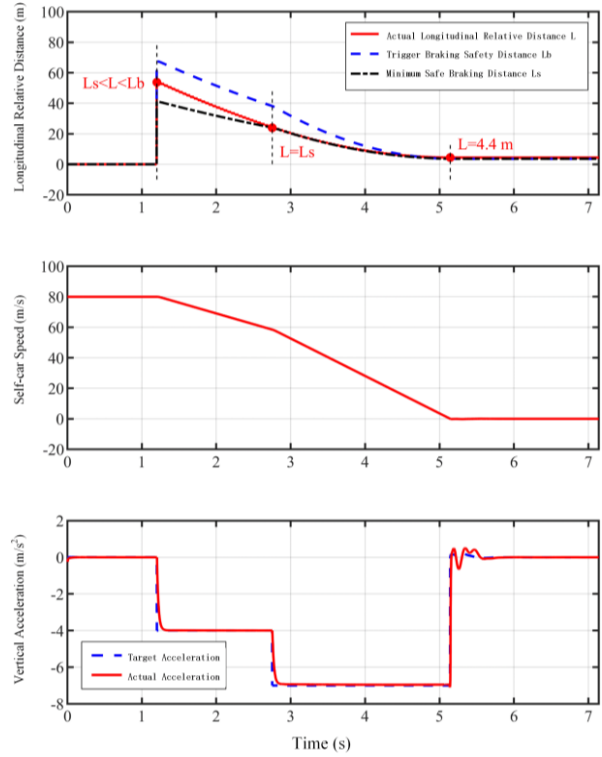

### 4.3 Collision Avoidance for Opposing Vehicles

The simulation results of the collision avoidance condition of the opposite-direction vehicle are shown in Figure 20. In the Step1 stage, the vehicle travels at a speed of 16.7 m/s. The obstacle vehicle crosses the centerline of the lane at 16.7 m/s and drives in the opposite direction of the self-propelled vehicle. At this time, the countdown of the collision time $TTC^{-1}$ is less than the set active steering collision avoidance threshold of 0.5, so the system does not perform active collision avoidance operation. In Step2 stage, the system detects that the countdown of the collision time $TTC^{-1}$ is greater than the set threshold, at this time the longitudinal relative distance between the self-vehicle and the obstacle vehicle is 66.7 m. The collision risk is considered to exist, but if the emergency braking collision avoidance operation is taken, the vehicle will still collide. At this time, the decision algorithm takes emergency steering to avoid the collision, and the system plans the collision avoidance path and carries out tracking. In the Step3 stage, the vehicle successfully crosses the obstacle vehicle coming from the opposite direction and continues the merging process. In the Step4 stage, the vehicle completes the collision avoidance process.

**Figure 20** Collision avoidance process for opposing vehicles.

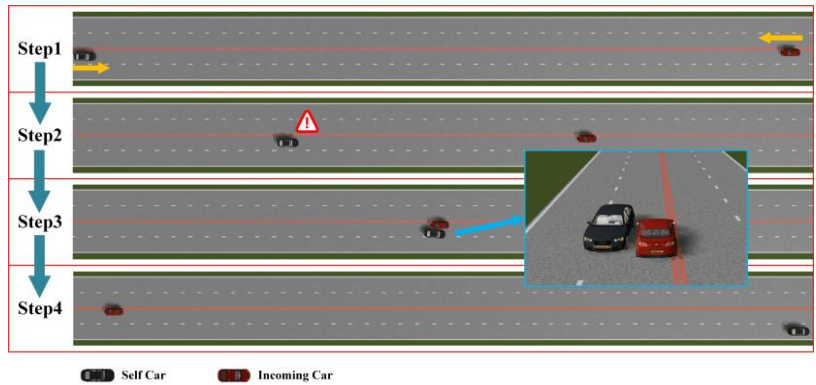

The results of collision avoidance path planning are shown in Figure 21:

**Figure 21** Planning path and constraint settings for collision avoidance for opposing vehicles.

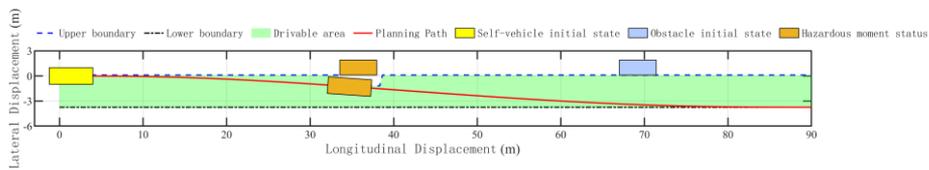

The tracking results of the vehicle for the collision avoidance path are shown in Figure 22. From the tracking result curve, it can be concluded that because the self-car has the judgment of collision risk in advance for the vehicles coming from the opposite direction, the time left for the collision avoidance phase (Step2) is also longer, and the planned path is also more gentle, the tracking lateral deviation is less than 0.1 m, and the heading deviation is basically kept within 0.005 rad, and the steering wheel changes smoothly. This also proves the collision avoidance capability of the system for the special situation of oncoming traffic.

**Figure 22** Tracking results of oncoming traffic steering collision avoidance path.

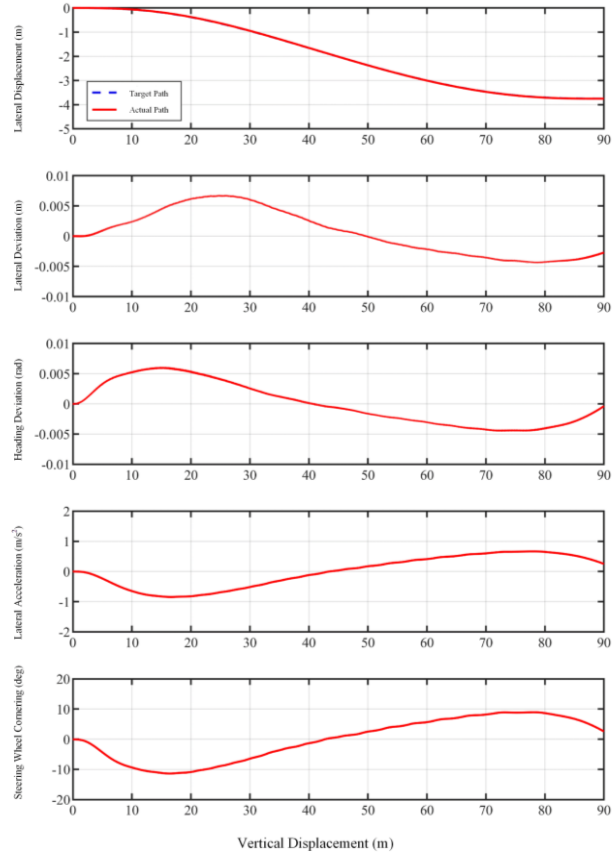

## 5   Conclusion

This paper proposes a multi-level decision framework obstacle avoidance algorithm for emergency scenarios. The steering system and braking system are integrated with this algorithm, and the collision avoidance operations such as collision warning, emergency braking, and emergency steering are taken according to different risk levels, which improves the lateral maneuverability of the system. When the system takes steering for obstacle avoidance, the planning algorithm can consider the vehicle's drivable area and generate the motion trajectory by solving the optimization problem with multiple constraints. The conclusions are drawn as follows:

(1) The decision avoidance algorithm can make correct operations quickly in the four typical hazardous traffic scenarios in this paper;

(2) The planning module can generate smooth collision-free trajectories suitable for lane-changing maneuvers in complex traffic situations, and the tracking control module can complete tracking control tasks accurately and stably;

(3) These results could motivate future work on incorporating dynamic prediction models for real traffic environments.